\documentclass[journal]{IEEEtran}
\ifCLASSINFOpdf
\else
\fi

\usepackage{color}
\usepackage{graphicx}
\usepackage{latexsym}
\usepackage{amsmath}
\usepackage{algorithm}
\usepackage{algorithmic}
\usepackage{etoolbox}
\makeatletter
\patchcmd{\@makecaption}
  {\scshape}
  {}
  {}
  {}
\makeatother

\begin{document}
%
\title{Method for Constructing Artificial Intelligence Player with Abstraction to Markov Decision Processes in Multiplayer Game of Mahjong}
%
%
%


\author{
    \IEEEauthorblockN{Moyuru Kurita\IEEEauthorrefmark{1} and Kunihito Hoki\IEEEauthorrefmark{2}} \\
    \IEEEauthorblockA{\IEEEauthorrefmark{1}Heroz Inc.}, 
    \IEEEauthorblockA{\IEEEauthorrefmark{2}The University of Electro-Communications}
    \thanks{Corresponding author: Moyuru Kurita (email: mkmjai1@gmail.com).}
}

\maketitle

\begin{abstract}
We propose a method for constructing artificial intelligence (AI) of mahjong, which is a multiplayer imperfect information game.
Since the size of the game tree is huge, constructing an expert-level AI player of mahjong is challenging.
We define multiple Markov decision processes (MDPs) as abstractions of mahjong to construct effective search trees.
We also introduce two methods of inferring state values of the original mahjong using these MDPs.
We evaluated the effectiveness of our method using gameplays vis-\`{a}-vis the current strongest AI player.
\end{abstract}


%
\IEEEpeerreviewmaketitle

\section{Introduction}
\IEEEPARstart{M}{ahjong} is a popular game in Asia and has been played over a hundred years with different rule sets according to the country or regions.
Most rule sets of mahjong share common properties that makes developing AI challenging, e.g., the number of players is three or four (mostly four), size of the game tree is huge, size and number of information sets are large, and uncertainty strongly influences gameplay.
Though the performance of AI has exceeded human experts in most two-player perfect information games and some multiplayer imperfect information games, this has not been the case for mahjong.

We propose a method of constructing an AI mahjong player and demonstrate that its performance is better than current AI players.
We abstract the game of mahjong and treat it as multiple Markov decision processes (MDPs).
We considered the averaged behavioral strategies of a variety of experts to replace three of the four players with a chance player.
We introduce four MDPs as abstractions of mahjong and formulate a value functions by using these MDPs.
The action probabilities of the chance player acting on behalf of three players are inferred from game records of experts and the authors' experience.
We also verified the performance of greedy players who always choose an action of the greatest value.


This paper is organized as follows.
We explain the rules and features of mahjong in Sec.~\ref{sec:rule}.
We review related research in Sec.~\ref{sec:previous-research}, and explain the contributions of our research in Sec.~\ref{sec:contribution}.
We briefly outline our method in Sec.~\ref{sec:method-abstract}, and give further details of it in Sec.~\ref{sec:method}.
We discuss the performance evaluation of our method using gameplays vis-\`{a}-vis existing the current strongest AI player in Sec.~\ref{sec:experiment}.

This research developed the contents of research on AI player of mahjong released at the domestic conference \cite{Kurita}, organized the theoretical framework of the method, added a new computer experiment, and summarized it newly.

%
%
%
%
 

\section{Rules and Features of Mahjong}
\label{sec:rule}

\subsection{Outline of Rules}
There are variations in the rules of mahjong, but this section outlines the most basic mahjong rules commonly used in Japan (see \cite{Miller}).
Mahjong is a game played by four people.
They use four sets of 34 tiles.
These 34 tiles are different, and the total number of tiles is 136.
Each player starts with 25,000 points.
One gameplay of mahjong is a sequence of multiple hands\footnote{A hand also means a set of tiles owned by a player}, and the points move from player to player by each hand.
A standard way to earn points is to form a winning hand earlier than the other players.
A typical wininning hand consists of four combinations of three tiles satisfying specific conditions (each combination is called mentsu) and one pair tiles of the same kind.
The final rank of each player is determined by the final points of the game.

In addition to the four players, we consider a chance player who introduces contingency into gameplay.
The actions of the chance player are classified into the following two types.
\begin{itemize}
\item $A_{\text{HandDistribution}}$:
The chance player distributes hands from the draw pile to each player.
Each player receives a hand composed of 13 tiles, and one player receives an additional tile as $A_{\text{Draw}}$.
This player is called the parent player.
\item $A_{\text{Draw}}$: The chance player distributes one tile from the draw pile to a player.
The tile is not revealed to the other players.
\end{itemize}

The information sets of each player are categorized into two types.
Any information set of the first type follows $A_{\text{HandDistribution}}$ or $A_{\text{Draw}}$.
The player dealt $A_{\text{Draw}}$ is the player to choose an action from one of the following action types.
\begin{itemize}
\item $a_{\text{DrawWin}}$:
The player declares a win when his/her hand (13 + 1 tiles) satisfies specific conditions.
Then the player discloses the hand and earns points depending on the hand from the other players.
All players then discard their hands.
\item $A_{\text{Discard}}$:
The player discards a tile (therefore, the size of the hand is kept at 13).
The tile is now revealed to the others.
\end{itemize}
The second type follows $A_{\text{Discard}}$ or $A_{\text{Take\&Discard}}$ (explained below) of a player $i$ who discarded tile $h$, where the other players $j$ sometimes gains the right to choose one of the following action types.
\begin{itemize}
\item  $a_{\text{TakeWin}}$:
Player $j$ declares a win when his/her hand (13 tiles) and $i$'s discarded $h$ (1 tile) satisfy specific conditions.
Then $j$ earns points from $i$, and all players discard their hands.
\item $A_{\text{Take\&Discard}}$:
Player $j$ assembles a mentsu using $h$ (take), discloses the mentsu, then discards another tile.
Take behaviors are classified into a few classes such as pon (also known as pung) and chi (also known as chow) depending on mentsus assembled.
\item $a_{\text{Pass}}$:
Player $j$ does not declare anything.
If all players pass, the next action is $A_{\text{Draw}}$ of a player next to $i$.
\end{itemize}

\begin{figure}[htb]
\centering
\includegraphics[bb=0 124 317 540,scale=0.6]{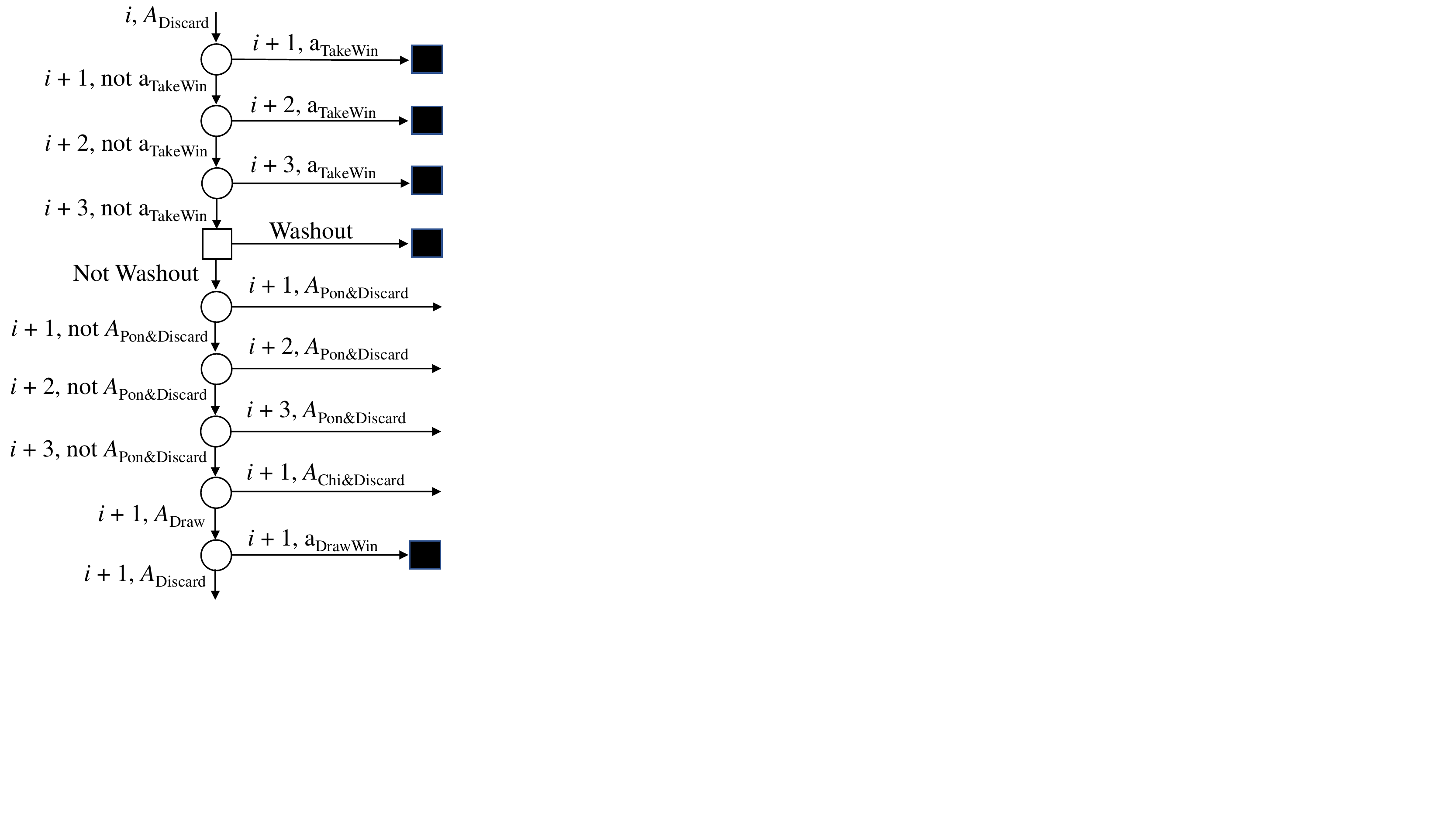}
\caption{
Some branches of hand.
White circles represent information sets in which players choose actions.
Black squares represent sets of endings of hand.
White square represents branch of washout determined by rules.
}
\label{fig:shinkou}
\end{figure}

These action types form the bulk of branching points in the hand.
Each hand starts with $A_{\text{HandDistribution}}$, and the hand ends when one of the players chooses an action of type $A_{\text{Win}}$ (set of $a_{\text{DrawWin}}$ and $a_{\text{TakeWin}}$), or when the number of tiles in the draw pile decreases to a specific number.
Fig.~\ref{fig:shinkou} illustrates some branches of the gameplay from $A_{\text{Discard}}$ of player $i$ to $A_{\text{Discard}}$ of player $i+1$ (player 5 means 1).
A hand consists of about 60 of such parts.

The rules specify whether the next hand starts or the entire game ends when a hand ends.
The chance player determines one parent from four players for the first hand, and one parent for each subsequent hand is specified by the rules.
If the rule called tonpu-match is applied, then a player usually plays four to six hands in a gameplay, and the player usually plays one or
more hands as a parent.

We now describe several important terms in mahjong that we use in this paper.
\begin{itemize}
\item tenpai:
When a hand (13 tiles) becomes a winning hand with one tile, the hand is called tenpai.
\item shanten-number:
The minimum number of tiles that need to be exchanged to make the hand tenpai.
\end{itemize}

\subsection{Features of Mahjong}
Mahjong's gameplay consists of playing multiple hands in a row.
The game situation before $A_{\text{HandDistribution}}$ can be explained from only a small amount of shared information $\phi_{\text{hand}}$ (points of four players etc.).
Also, four behavior strategies and the shared information determine the expected value of the final ranking of each player.
Since it is possible to represent the game situation before $A_{\text{HandDistribution}}$ by $\phi_{\text{hand}}$ and obtain a sufficient number of expert's record, the final ranking in this game situation is easily predicted by regression.
Therefore, it is reasonable to represent a hand as a truncated partial game.
The game tree handles the end of the hand (i.e., the beginning of the next hand) as terminal nodes to which the expected values of the final rank are given.
Similar methods of treating the entire game as a continuous truncated partial game are used in other games.
For example, it is common to play one game of $n$ point match backgammon as an individual game based on the reward of the match equity table \cite{Woolsey}.

Let $P(\text{RANK}(x,i) | \phi_{\text{hand}})$ be the
probability that event $\text{RANK}(x,i)$, i.e., player $i$ acquires rank $x$ for $i,x \in \{1,2,3,4\}$, occurs under the condition $\phi_{\text{hand}}$.
Then the expected value of player $i$'s payoff at $\phi_{\text{hand}}$ is given by
\begin{eqnarray}
U^{i}_{\text{hand}}(\phi_{\text{hand}}) = \sum_{x \in \{1,2,3,4\} } P(\text{RANK}(x,i) | \phi_{\text{hand}})U_{\text{game}}(x).
\label{eq:hand-end}
\end{eqnarray}
Here, $U_{\text{game}}(x^{i}_{\text{rank}})$ is the payoff of rank $x$, which is defined by the rules of the tournament (normally, the higher the rank, the higher the payoff).


We roughly estimate the number of information sets (i.e., decision points) of a player in a truncated partial game of one hand by ignoring $a_{\text{TakeWin}}$, $A_{\text{Take\&Discard}}$, and $a_{\text{Pass}}$.
There are about $10^{11}$ ways to distribute hands to a player by $A_{\text{HandDistribution}}$.
The number of legal actions in $A_{\text{Discard}}$ is about ten.
After that, the player can see about 30 kinds of tiles at each $A_{\text{Discard}}$ of the other three players.
Then, the player can see about 30 kinds of tiles at $A_{\text{Draw}}$.
We call a partial gameplay from discard to next draw of the same player a turn.
Since the number of turns to play one hand is about 20 at most, we obtain a rough estimate by
\begin{eqnarray}
10^{11} \times (30^4 \times 10)^{20} \approx 10^{150}.
\end{eqnarray}
The exponent value is a little smaller than that of the Go state space \cite{Allis}.

A hand falls into five scenarios from player $i$'s point of view.
\begin{enumerate}
\item win:
$i$ chooses an action in  $A_{\text{Win}}$.
Usually this is the most favorable scenario.
\item lose:
Another player chooses $a_{\text{TakeWin}}$ against a tile discarded by $i$.
Usually this is the most unfavorable scenario.
\item other win:
Another player chooses $a_{\text{DrawWin}}$ or $a_{\text{TakeWin}}$ against a tile discarded by another player
different from $i$.
It is difficult to realize this scenario with $i$'s will.
\item tenpai washout:
The hand ends due to a shortage of the draw pile when $i$ has a tenpai hand.
\item noten washout:
The hand ends due to a shortage of the draw pile when $i$ does not have a tenpai hand.
\end{enumerate}
We ignore other scenarios because they are rare.
Choosing one of these scenarios according to the current game situation is one of the most important strategies for playing mahjong.

\section{Previous Research}
\label{sec:previous-research}

Due to research over the past 20 years, AI has exceeded human ability in many two-player zero-sum games with perfect information, e.g., backgammon~\cite{Tesauro}, checkers~\cite{Schaeffer1}, chess~\cite{Campbell}, shogi~\cite{Hoki, Silver1140}, and Go~\cite{Silver}.
One of the techniques that has played a central role in the development of these AI players is heuristic search using the property that two players share symmetric information \cite{Russell}.
However, heuristic search has not been powerful in games with three or more players and imperfect information.
The reason for this is that it is difficult to construct a search tree that is easy to finish searching and effective for representing proper game situations.

There are also interesting research results from two-player games with imperfect information.
Counterfactual regret minimization (CFR) is a powerful technique based on self-play for constructing a strong player of a game belonging to such a class \cite{Zinkevich}.
In fact, $\epsilon$-Nash equilibrium of heads-up limit Texas hold'em, which has about $10^{14}$ decision points for a player, was obtained using $\text{CFR}^{+}$, a variant of CFR \cite{Bowling}.
Moreover, an expert-level AI player of heads-up no-limit Texas hold'em, which has more than $10^{160}$ decision points, has been developed using tree search with bet abstraction and deep learning of counterfactual values \cite{Moravcik}.
In research other than on poker AI, an expert-level AI of Scrabble has been developed using a selective move generator, simulations of likely game scenarios, and the heuristic search algorithm $B^{*}$\cite{Sheppard}.

Relatively few studies have been reported on multiplayer imperfect-information games such as mahjong.
Even in such games, one of the research objectives may also be to compute approximations to some of Nash equilibrium points.
A case study on limit Texas hold'em with three players was conducted \cite{Risk} in which an AI player based on CFR outperformed other AI players, although this method loses the theoretical guarantees of two-player zero-sum games.
However, applying CFR variants to other multiplayer games is not easy.
Implementation of a mahjong player based on CFR is difficult because the size of the game tree is too large to search, and the abstraction for reducing the search space is unknown.

Another research objective in multiplayer imperfect-information games is to construct an AI player by using heuristic methods, which are known to be effective in two-player perfect-information games.
There are AI players in multiplayer Texas hold'em.
Poki, which is an AI player of Texas hold'em with multiple players, adopts a betting strategy based on heuristic evaluation of hand strength \cite{Billings}.
Commercial software called Snowie is considered to have the same strength as experts, but its algorithm is unpublished.

Besides poker games, an expert-level AI player of Skat has been constructed based on heuristic search algorithms of perfect-information games.
The search algorithms have been used in the game using game-state inference and static evaluation obtained by regression using game records \cite{Buro}.
It is interesting to build AI players based on such heuristic search algorithms in other games with multiplayers and imperfect information, but it is difficult to construct an effective search tree.
In fact, it has been reported that an AI player of The Settlers of Catan applying Monte-Carlo tree-search methods is not as strong as human players \cite{Szita}.

There has been research on AI players of mahjong.
There is an open-source beginner-level player based on the Monte-Carlo simulation called manue\footnote{Hiroshi Ichikawa https://github.com/gimite/mjai-manue}.
To model actions of opponent players statically, it uses inferred probabilities that an action in $A_{\text{Discards}}$ (sum of $A_{\text{Discard}}$ and $A_{\text{Take\&Discards}}$) by a player induces a win of another player.
Bakuuchi is another player that carries out Monte-Carlo simulations.
Early Bakuuchi uses such probabilities with higher accuracy, Eq. (\ref{eq:hand-end}), to evaluate each simulation at the end of the hand and simulation policies learned from game records \cite{Mizukami}.
In that study, they reported that point dependency on the policy is inappropriate and had reached only the intermediate level.
Note that recent Bakuuchi, which is unpublished, has reached the advanced level.
To the best of our knowledge, no tree has yet been discovered to search for better decisions.

Our method abstracts mahjong to construct effective search trees to appropriately deal with various game situations.
Game abstraction is known as an effective means to reduce a huge search space of an extensive-form game with imperfect information \cite{Sandholm}.
For example, the effectiveness of information and action abstraction is shown in the aforementioned poker and patrolling security games \cite{Basilico}.

\section{Contributions}
\label{sec:contribution}



The contribution of this paper are as follows.

(1) We define an abstraction of mahjong, \textit{Inclusive Policy Solitary Mahjong} $\cal{M}$. $\cal{M}$ is an MDP that is expected to be effective to evaluate a short-term behavior strategies to compete on the most favorable scenario win.
Three other players are replaced with a static environment, and the decision-making player goes through the process $\cal{M}$ and ends with a win, lose, other win, tenpai washout, or noten washout scenario.

(2) We introduce several features in machine learning that are expected to be representative of a long-term behavior strategies of a hand and be useful for inferring state values.
The features are computed using three other MDPs.
Three other players are replaced with static environment, and the decision-making player goes through each process and ends with a few specific scenarios.

(3) We propose a method for constructing an AI player using (1) and (2).
We present the experimental results of 3557 gameplays with the state-of-the-art AI mahjong player, in which our AI player achieved significantly higher average rank
We also present that our player makes each decision in a few seconds using a realistic computational resource.

\section{Outline of Proposed Method}
\label{sec:method-abstract}

We discuss action values of mahjong by separating the cases in which a hand ends immediately.
Let us consider the first few actions from information set $u_0$.
Recall that most actions belong to three types, $A_{\text{Win}}$,  $A_{\text{Discards}}$, and $a_{\text{Pass}}$.

We first consider action type $A_{\text{Win}}$.
After such an action, $a$, a hand ends without any action of the other players.
When player $i$ at $u_0$ takes $a$, the action value is
\begin{eqnarray}
Q^{\text{org}}(u_{0}, a) = U^{i}_{\text{hand}}(\phi_{\text{hand}}).
\end{eqnarray}
We compute $U^i_\text{hand}(\phi_\text{hand})$ using Eq.(\ref{eq:hand-end}), where $P(\text{RANK}(x,i) | \phi_{\text{hand}})$ is inferred using a multi-class logistic regression model, as in a previous study \cite{Mizukami}.

Next, we consider action type $A_{\text{Discards}}$.
Such an action, $a$, is accompanied by discarding a tile, and the hand also ends immediately if another player chooses $a_{\text{TakeWin}}$ against the tile.
Let us assume that the other players determine actions according to static probability and treat them as if they are also the chance player.
When $i$ at $u_0$ takes $a$, we approximate the action value as
\begin{eqnarray}
Q^{\text{org}}(u_{0}, a) &\approx& P(a_{\text{TakeWin}} \text{ from} \ i | u_{0}, a) U_{a_{\text{TakeWin}} \text{ from} \ i}(u_{0}, a) \notag \\
&+& P(\overline{a_{\text{TakeWin}} \text{ from} \ i} | u_{0}, a) U_{\overline{a_{\text{TakeWin}} \text{ from} \ i}}(u_{0}, a). \notag \\
\label{eq:value-dahai}
\end{eqnarray}
The probability $P(a_{\text{TakeWin}} \text{ from} \ i | u_{0}, a)$ and corresponding expected payoff  $U_{a_{\text{TakeWin}} \text{ from} \ i}(u_{0}, a)$ can be inferred using orthodox machine learning methods because a hand immediately terminates if $a$ is followed by the $a_{\text{TakeWin}}$ of another player.
We discuss these methods in Sec.~\ref{subsec:heuristic}.

We then consider action type $a_{\text{Pass}}$.
When $i$ at $u_0$ takes such an action, $a$, we approximate the action value as
\begin{eqnarray}
Q^{\text{org}}(u_{0}, a) \approx U_{a_{\text{Pass}}}(u_{0}, a).
\label{eq:value-pass}
\end{eqnarray}
The value $U_{a_{\text{Pass}}}(u_0,a)$ is the corresponding expected payoff.

After separating the cases of immediate ends of a hand, we need to compute $U_{\overline{a_{\text{TakeWin}} \text{ from} \ i}}(u_{0}, a)$ and $U_{a_{\text{Pass}}}(u_{0}, a)$ to estimate the action value at $u_0$.
Our method uses two models to compute these values.
These models represent the game state $s$, which can be determined from $(u_0, a)$ to the end of a hand, as tuple $s=(u_0, q, h, c, t)$.
Here, $q$ is $i$'s hand, $h$ is a tile obtained by $i$ most recently, $c$ is a type of state described below, and $t$ is the number of tiles discarded by $i$ since $u_0$. We omit $u_0$ of $s$ below.

To set up the first model, in addition to using the state representation, we define inclusive policy one-player mahjong $\cal{M}$ which is an MDP and takes into account as many scenarios as possible.
This MDP requires comprehensive search and is designed to predict hands ending with a relatively small number of steps with high accuracy.

To set up the second model, in addition to using the state representation, we define several one-player mahjong games, which are different MDPs, and take into account different small subsets of all scenarios.
The estimation of action values by one of these one-player mahjong games is not accurate because each subset is restricted.
However, these one-player mahjong games are amenable to long-term computation and can be used to provide good features to predict the scenario of hands with a relatively large number of steps.

\section{Proposed Method}
\label{sec:method}

This chapter is organized as follows.
In Sec.~\ref{subsec:MDP}, we define multiple MDPs as mahjong abstractions and formulate their action-value functions.
Then we represent action values of the original game from these MDPs in Sec.~\ref{subsec:value_inference}.
In Sec.~\ref{subsec:heuristic}, we describe methods of calculating input parameters of the MDPs.
In Sec.~\ref{subsec:search}, we describe an efficient search algorithm of the MDPs.

\subsection{Abstraction to MDPs}
\label{subsec:MDP}
Consider player $i$ at information set $u_0$ of a hand which is a truncated partial game of mahjong.
We abstract the hand rooted at $u_0$ to an MDP in four ways.
Here, $i$ is the agent who makes decisions, and decision making of the others are represented by transitions probabilities of states.
This section defines four MDPs and formulas that approximately represent the expected value the final ranking of $i$.

\subsubsection{Inclusive Policy Solitary Mahjong $\cal{M}$}
MDP $\cal{M}$ covers various scenarios from $i$'s point of view.
Type $c$ of state $s$ in $\cal{M}$ indicates one of the following:
\begin{itemize}
\item $S_{\text{Discard}}$:
Player $i$ at $s$ of this type can choose $a_{\text{DrawWin}}$ to gain payoff $U_{\text{DrawWin}}(q,h)$ only if $q$ and $h$ satisfy conditions used in the original game.
If $i$ does not, then $i$ has to choose an action in $A_\text{Discard}$ to discard a tile from $q$ and $h$.
\item $S_{\text{Take}}$:
Player $i$ at $s$ of this type can choose $a_{\text{TakeWin}}$ to gain payoff $U_{\text{TakeWin}}(q,h)$ or an action in $A_{\text{Take\&Discard}}$ only if $q$ and $h$ satisfy conditions used in the original game.
If $i$ does not, then $i$ chooses $a_{\text{Pass}}$.
\item $S_{\text{Fold}}$:
Player $i$ at $s$ of this type chooses either $a_{\text{Fold}}$ or $a_{\text{NotFold}}$.
If $i$ chooses $a_{\text{Fold}}$, $i$ gains payoff $U_{\text{Fold}}(q,t)$.
\end{itemize}
$S_\text{Discard}$ and $S_\text{Take}$ correspond to $i$'s information sets following $A_\text{Draw}$ or $A_\text{Discard}$ of other players in the original game.
Though an information set corresponds to $S_{\text{Fold}}$ does not exist in the original game, we introduce this type of states for simplification.

\begin{figure}[htb]
\centering
\includegraphics[scale=0.5]{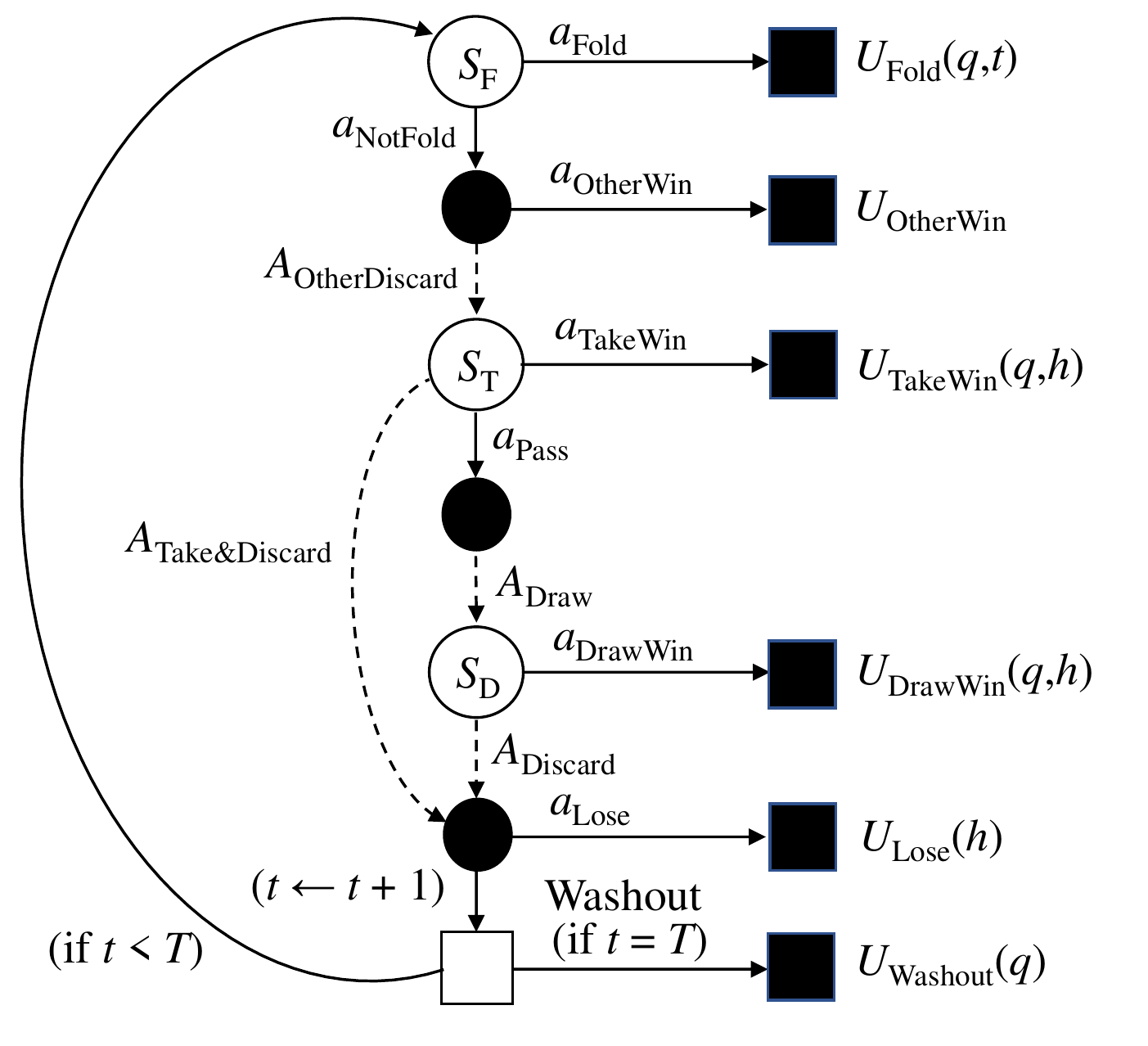}
\caption{
Some branches of hand.
White circles represent information sets in which players choose actions.
Black squares represent sets of endings of hand.
White square represents branch of washout determined by rules.
}
\label{fig:inclusive}
\end{figure}

MDP $\cal{M}$ terminates immediately if $i$ chooses either $a_{\text{DrawWin}}$, $a_{\text{TakeWin}}$, or $a_{\text{Fold}}$.
Otherwise, the chance player choose actions, which are categorized as follows. 
\begin{itemize}
\item $a_{\text{Lose}}$:
$\cal{M}$ terminates at probability $P(a_{\text{Lose}}|q,h,t)$ after $i$'s action of $A_{\text{Discard}}$ or $A_{\text{Take\&Discard}}$, where $h$ is a tile discarded by the action and $i$ gains payoff $U_{\text{Lose}}(h)$.
If $\cal{M}$ does not terminate, the action number $t$ increases by one.
Then $\cal{M}$ terminates if $t=T$, and $i$ gains payoff $U_{\text{washout}}(q)$.
Otherwise, the state transfers to a state of $S_{\text{Fold}}$.
($T$ is an input parameter of $\cal{M}$, which will be described in Sec.\ref{subsec:heuristic})
\item $a_{\text{OtherWin}}$:
$\cal{M}$ terminates at probability $P(a_{\text{OtherWin}}|q,t)$ after $i$ choose $a_{\text{NotFold}}$, and $i$ gains payoff $U_{\text{OtherWin}}$.
Otherwise, the chance player choose an action of $A_{\text{OtherDiscard}}$.
\item $A_{\text{OtherDiscard}}$:
The chance player chooses tile $h$ at probability $P_{\text{T}}(h|q,t)$ and the state transfers to a state of $S_{\text{Take}}$.
\item $A_{\text{Draw}}$:
The chance player deals tile $h$ at probability $P_{D}(h | q,t)$ after $i$ chooses $a_{\text{Pass}}$ at a state of $S_{\text{Take}}$, and the state transfers to a state of $S_{\text{Discard}}$.
\end{itemize}
$A_{\text{Draw}}$ corresponds to that of the original game, while other types of branches correspond to the averaged actions of other players of the original game.
$a_{\text{Lose}}$ and $a_{\text{OtherWin}}$ correspond to lose and other win scenarios in the original game, respectively.
The flow of $\cal{M}$ is schematically shown in Fig.~\ref{fig:inclusive}.

When the type of state is $S_\text{Discard}$, i.e., $s_{\text{D}} = (q,h, S_\text{Discard},t)$, the action-value function is as follows.
For $(q, h)$, where $a_\text{DrawWin}$ is legal, we have
\begin{eqnarray}
&&Q(s_{\text{D}}, a_\text{DrawWin}) = U_{\text{DrawWin}}(q,h).
\end{eqnarray}
MDP $\cal{M}$ terminates with this action.
When action $a$ is in $A_\text{Discard}$ and tile $h'$ is selected, we have
\begin{eqnarray}
Q(s_{\text{D}}, a) &=& P(a_\text{Lose}|q',h',t) U_{\text{Lose}}(h') \notag \\
&+& P(\overline{a_\text{Lose}}|q',h',t) V_{\overline{a_\text{Lose}}}(q',h',t+1). \notag \\
V_{\overline{a_\text{Lose}}}(q',h',t+1) &=& \begin{cases}
V(s'_{\text{F}}) & t + 1 < T \\
U_{\text{washout}}(q') & t + 1 = T
\end{cases}
\label{eq:exp_ext_playerD}
\end{eqnarray}
Here, $s'_{\text{F}} = (q', \text{null}, S_\text{Fold}, t+1)$ and $q'$ is the hand after discarding $h'$ from $q$ and $h$.

When the type of state is $S_\text{Fold}$, i.e., $s_{\text{F}} = (q, \text{null}, S_\text{Fold} ,t)$, and the action is $a_\text{NotFold}$, we have
\begin{eqnarray}
Q(s_{\text{F}}, a_\text{NotFold}) &=& P(a_{\text{OtherWin}} | q,t) U_{\text{OtherWin}} \notag \\
&+& P(\overline{a_{\text{OtherWin}}} | q,t) \sum_{h \in H} P_{T}(h | q,t)V(s_{\text{T}}), \notag \\
\end{eqnarray}
where $s_{\text{T}} = (q, h, S_{\text{Take}}, t)$ and $H$ is the set of all tile kinds.

When the type of state is $S_\text{Take}$, i.e., $s_{\text{T}} = (q,h,S_\text{Take},t)$, and the action is $a_\text{Pass}$, we have
\begin{eqnarray}
Q(s_{\text{T}}, a_\text{Pass}) &=& \sum_{h' \in H} P_{D}(h' | q,t)V(s'_{\text{D}}),
\end{eqnarray} 
where $s'_{\text{D}} = (q, h', S_\text{Discard},t)$.
When action $a$ is in $A_\text{Take\&Discard}$, we have
\begin{eqnarray}
Q(s_{\text{T}}, a) &=& P(a_\text{Lose}|q', h', t) U_{\text{Lose}}(h') \notag \\
&+& P(\overline{a_\text{Lose}}| q', h', t)  V_{\overline{a_\text{Lose}}}(q',h',t+1),
\end{eqnarray}
where $h'$ is the discarded tile.
The formulas of payoff functions are outlined in Sec. \ref{subsec:heuristic}.

\subsubsection{Folding Solitary Mahjong ($\cal{M}_{\text{fold}}$)}
\label{sec:fold-policy}
MDP $\cal{M}_{\text{fold}}$ covers two scenarios (lose and other win) to represent folding strategies of player $i$.
Folding is a behavior strategy in which $i$ abandons the most favorable scenario win and avoids the most unfavorable scenario lose of the current hand.
Actions types $A_{\text{Draw}}$ and $A_{\text{Discards}}$ of the other players are ignored to simplify the game.

In $\cal{M}_{\text{fold}}$, under the probability of lose $P_{\text{Lose}}(h)$ and payoff of lose $U_{\text{Lose}}(h)$, $i$ is only allowed to discard $h$ from $i$'s hand.
The action type of $i$ is only $A_{\text{Discard}}$, the state type of $i$ is only $S_{\text{Discard}}$, and the number of tiles in $i$'s hand decreases monotonically from the initial state because actions in $A_{\text{Draw}}$ are ignored.
There are two types of branches due to the chance player as follows.
\begin{itemize}
\item $a_{\text{Lose}}$:
$\cal{M}_{\text{Fold}}$ terminates with probability $P_{\text{Lose}}(h)$ of $h$ discarded just before, and $i$ gains payoff $U_{\text{Lose}}(h)$. If $i$ have discarded $h$ more than once from the initial state, this type is not selected.
\item $a_{\text{OtherWin}}$:
When $\cal{M}_{\text{fold}}$ does not terminate by $a_{\text{Lose}}$, it terminates with constant probability $\alpha$ (we tentatively set $\alpha = 0.1$) and $i$ gains payoff $U_{\text{OtherWin}}$.
Otherwise, the state transfers to $S_\text{Discard}$.
\end{itemize}
MDP $\cal{M}_{\text{fold}}$ also terminates when $i$ discards all or $T$ tiles and $i$ gains payoff $U_{\text{OtherWin}}$.
For the sake of simplicity, we assume a natural condition $U_{\text{Lose}}(h) < U_{\text{OtherWin}}$ holds for all $h$.
Under this assumption, if $i$ has tiles that have been discarded once or more, $i$ should always discard one of these tiles.

Let $q$ be $i$'s hand, $h$ be a tile kind to discard, and $t$ be the number of discards from the initial state.
Action-value functions are formulated in terms of $(q, t)$, which specifies a state in $S_{\text{Discard}}$, and $h$, which specifies an action in $A_{\text{Discard}}$ as
\begin{eqnarray}
Q(q,t,h) &=& \begin{cases}
Q_{\text{new}}(q, t, h) & \text{if $h$ is a new discard} \\
Q_{\text{prev}}(q, t, h) & \text{otherwise}
\end{cases} \notag \\
Q_{\text{new}}(q, t, h) &=& P_{\text{Lose}}(h)U_{\text{Lose}}(h) \notag \\
&+& (1-P_{\text{Lose}}(h)) \alpha U_{\text{OtherWin}} \notag \\
&+& (1-P_{\text{Lose}}(h)) \bar{\alpha}V(q',t+1) \notag \\
Q_{\text{prev}}(q, t, h) &=& \alpha U_{\text{OtherWin}}+ \bar{\alpha}V(q',t+1).
\label{eq:MDP-fold}
\end{eqnarray}
Here, $\bar{\alpha} = 1 - \alpha$, and $q'$ is a hand where $h$ is subtracted from $q$.

The optimal policy is to discard the tiles in ascending order of $f(h)$, i.e.,
\begin{eqnarray}
f(h) &=& \frac{ P_{\text{Lose}}(h)(U_{\text{OtherWin}} - U_{\text{Lose}}(h)) }{ 1 - (1- P_{\text{Lose}}(h)) \bar{\alpha}^{n_{h}} },
\label{eq:fold-order}
\end{eqnarray}
where $n_{h}$ is the number of tiles of kind $h$ in $i$'s hand in the initial state, and the optimal value is given by
\begin{eqnarray}
E_{\text{Fold}} &=& P_{\text{Lose}}(h_{1})U_{\text{Lose}}(h_{1}) \notag \\
&+& (1-P_{\text{Lose}}(h_{1}))\left[ 1-\bar{\alpha}^{n_{h_{1}}} \right] U_{\text{OtherWin}} \notag \\
&+& (1-P_{\text{Lose}}(h_{1}))\bar{\alpha}^{n_{h_{1}}}P_{\text{Lose}}(h_{2})U_{\text{Lose}}(h_{2}) \notag \\
&+& \cdots \notag \\
&+& \left[ \prod_{k=1}^{K}(1-P_{\text{Lose}}(h_{k})) \right]\bar{\alpha}^{\sum_{k}n_{h_{k}}} U_{\text{OtherWin}}.
\label{eq:fold-series}
\end{eqnarray}
Here, $h_{k} \ (k=1, \cdots, K)$ is in ascending order of Eq.(\ref{eq:fold-order}), and $K$ is the number of tile kinds in $i$'s hand in the initial state.
The optimal policy ends up with the scenario lose with the probability
\begin{eqnarray}
P_{\text{FoldLose}} &=& P_{\text{Lose}}(h_{1}) \notag \\
&+& (1-P_{\text{Lose}}(h_{1}))\bar{\alpha}^{n_{h_{1}}}P_{\text{Lose}}(h_{2}) + \cdots.
\label{eq:ori_optimum_series}
\end{eqnarray}
Let $U_{\text{LoseAverage}}$ be the optimal value under the condition of lose termination, which is discussed in Sec. \ref{subsec:value_inference}.
Eq.(\ref{eq:fold-series}) can be transformed using $U_{\text{LoseAverage}}$ as follows
\begin{eqnarray}
E_{\text{Fold}} &=& P_{\text{FoldLose}} U_{\text{LoseAverage}} \notag \\
&+& ( 1 - P_{\text{FoldLose}}) U_{\text{OtherWin}}.
\label{eq:lose-average}
\end{eqnarray}


\subsubsection{Winning Solitary Mahjong ($\cal{M}_{\text{win}}$) and $Tenpai$ Solitary Mahjong ($\cal{M}_{\text{tenpai}}$)}
MDPs $\cal{M}_\text{win}$ and $\cal{M}_\text{tenpai}$ are specialized for representing win and tenpai strategies, respectively.
Both are expected to have a smaller search space than $\cal{M}$.
Terminal nodes that do not have direct relations to the purpose of win for $\cal{M}_\text{win}$ or tenpai for $\cal{M}_\text{tenpai}$ are ignored.
Specifically, terminal nodes related to $a_{\text{Lose}}$, $a_{\text{Fold}}$, and $a_{\text{OtherWin}}$ are ignored in both MDPs.
Moreover, the payoff of washout $U_{\text{washout}}(q)$ in $\cal{M}_{\text{win}}$ does not depend on $q$, which we write as $U_{\text{NotWin}}$.
Also, $i$ of $\cal{M}_{\text{tenpai}}$ is unable to take an action in $A_{\text{Win}}$.
We omit formulas of action values, but they are derived by replacing zero with probabilities of those actions.

\subsection{Value Inference Using Multiple MDPs}
\label{subsec:value_inference}
In this section, we introduce two methods of inferring values of legal actions of the original game using multiple MDPs introduced in the previous section.
The first method simply adopts the optimal value $V^{*}$ of $\cal{M}$ to calculate the approximate values in Eqs. (\ref{eq:value-dahai}) and (\ref{eq:value-pass}) as
\begin{eqnarray}
U_{\overline{a_{\text{TakeWin}} \text{ from} \ i}}(u_{0}, a) &=& V^{*}(q, \text{null}, S_{\text{Fold}}, 1) \notag \\
U_{a_{\text{Pass}}}(u_{0}, a) &=& V^{*}(q, \text{null}, S_{\text{Fold}}, 0)
\label{eq:value-inference1}
\end{eqnarray}
where $u_0$ is player $i$'s information set and $q$ is $i$'s hand after action $a$.

The second method uses the results of value evaluations using $\cal{M}_{\text{win}}$, $\cal{M}_{\text{tenpai}}$, and $\cal{M}_{\text{fold}}$.
Let $Z$ be a set of hand scenarios $\{ \mathrm{win}, \mathrm{lose}, \mathrm{other}, \mathrm{tenpai}, \mathrm{noten} \}$.
This method calculates the approximate values in Eqs. (\ref{eq:value-dahai}) and (\ref{eq:value-pass}) as
\begin{eqnarray}
U_{\overline{a_{\text{TakeWin}} \text{ from} \ i}}(u_{0}, a) &=& V(q, 1) \notag \\
U_{a_{\text{Pass}}}(u_{0}, a) &=& V(q, 0) \notag \\
V(q,t) &=& \sum_{z \in Z} P(q,t,z) U(q,t,z)
\label{eq:state_value_ml}
\end{eqnarray}
We calculate $P(q, t, z)$ in Eq. (\ref{eq:state_value_ml}) using the product of probabilities obtained by playing these MDPs starting from initial state $(q, t)$.
The relations between $P(q,t,z)$ and these probabilities $p_\text{win}$, $p_{\text{washout}}$, $p_{\text{tenpai}}$, and $p_\text{lose}$ are
\begin{eqnarray}
P(q,t,\mathrm{win}) &=& p_{\text{win}}(q,t) \notag \\
P(q,t,\mathrm{tenpai}) &=& p_{\overline{\text{win}}}(q,t) p_{\text{washout}}(q,t) p_{\text{tenpai}}(q,t) \notag \\
P(q,t,\mathrm{noten}) &=& p_{\overline{\text{win}}}(q,t) p_{\text{washout}}(q,t) p_{\overline{\text{tenpai}}}(q,t) \notag \\
P(q,t,\mathrm{lose}) &=& p_{\overline{\text{win}}}(q,t) p_{\overline{\text{washout}}}(q,t) p_{\text{lose}}(q,t) \notag \\
P(q,t,\mathrm{other}) &=& p_{\overline{\text{win}}}(q,t) p_{\overline{\text{washout}}}(q,t) p_{\overline{\text{lose}}}(q,t)
\end{eqnarray}
These probabilities are inferred by logistic regression using features that are the results of value evaluations of these MDPs.
To explain their features, let us introduce the following symbols:
$V_{\text{win}}(q,t)$ and $P_{\text{win}}(q,t)$ are values from $\cal{M}_{\text{win}}$, where the former is a state value of $(q, \text{null}, S_{\text{Fold}}, t)$ and the latter is the probability that $i$ in this state finally chooses an action in $A_{\text{Wins}}$;
$P_{\text{tenpai}}(q,t)$ is the probability that $i$ in $(q, \text{null}, S_{\text{Fold}}, t)$ of $\cal{M}_{\text{tenpai}}$ will have a tenpai hand when it terminates;
and $P_{\text{Lose}}(q,t)$ and $U_{\text{LoseAverage}}(q,t)$ are values from Eqs. (\ref{eq:ori_optimum_series}) and (\ref{eq:lose-average}), where the initial hand of $\cal{M}_{\text{fold}}$ is $q$ and $T$ is adjusted according to $t$.
The features used for the regressions are as follows.
\begin{itemize}
\item $p_{\text{win}}(q,t)$:
\begin{itemize}
\item $\mathrm{logit}( P_{\text{win}}(q,t) )$
\item The number of players declaring riich (riich is discussed in Sec.~\ref{subsec:other}).
\item $1 - \prod_{j}(1-p^{j}_{\text{tenpai}})$. Here, $j$ runs over all players who is not $i$ and does not declaring riich.
\end{itemize}
\item $p_{\text{washout}}$:
\begin{itemize}
\item The number of players declaring riich.
\item $1 - \prod_{j}(1-p^{j}_{\text{tenpai}})$. Here, $j$ runs over all players who is not $i$ and does not declaring riich.
\end{itemize}
\item $p_{\text{tenpai}}(q,t)$:
\begin{itemize}
\item $\mathrm{logit}( P_{\text{tenpai}}(q,t) )$
\item The number of players declaring riich.
\end{itemize}
\item $p_{\text{lose}}(q,t)$:
\begin{itemize}
\item $\mathrm{logit}( P_{\text{Lose}}(q,t) )$
\item The number of actions in $A_{\text{Take\&Discard}}$ $i$ has chosen since $A_{\text{HandDistribution}}$.
\end{itemize}
\end{itemize}
Here, $p^{j}_{\text{tenpai}}$ is an inferred probability that player $j$ is tenpai at $u_0$.
This probability is modeled using logistic regression similar to that in a previous study~\cite{Mizukami}, but the difference is that the model is fitted for each number of $j$'s past actions in $A_{\text{Take\&Discard}}$ and for each number of $j$'s past actions in $A_{\text{Discards}}$ since $A_{\text{HandDistribution}}$.

We calculate $U(q, t, z)$ in Eq. (\ref{eq:state_value_ml}) as
\begin{eqnarray}
U(q,t,\mathrm{win}) &=& \frac{V_{\text{win}}(q,t) - P_{\text{win}}(q,t) U_{\text{NotWin}}}{P_{\text{win}}(q,t)} \notag \\
U(q,t,\mathrm{lose}) &=& U_{\mathrm{LoseAverage}}(q,t) \notag \\
U(q,t,\mathrm{other}) &=& U_{\mathrm{OtherWin}}.
\end{eqnarray}
We calculate $U(q,t,\text{tenpai})$ and $U(q,t,\text{noten})$ on the basis of mahjong rules and tenpai probabilities of the other players.
These probabilities, which should be those when a hand ends strictly speaking, are inferred at $u_0$.

\subsection{Parameters Used in MDPs}
\label{subsec:heuristic}
This section describes methods for determining parameters in the MDPs.
Let the agent of these MDPs be player $i$, and $i$'s current information set of the original mahjong be $u_0$ as before.
The first parameter to be described is $T$.
Let  $T_{\text{max}}$ be the maximum number of $i$'s future actions in $A_{\text{Discards}}$ until the current hand ends assuming that no player will choose actions in $A_{\text{Take\&Discard}}$.
We set $T$ to $T_{\text{max}}$ for $\cal{M}$.
For the other MDPs, we set $T$ to $\lceil T_{\text{max}}\sigma_{\text{ratio}} \rceil$.
We determine ratio $\sigma_{\text{ratio}}$ on the basis of logistic regression using the same features as those used for $p_{\text{washout}}$ and label
$T_{\text{measured}}/T_{\text{max}}$, where $T_{\text{measured}}$ is the number of future actions in $A_{\text{Discards}}$ until the current hand ends.
The training data (the pairs of features and a label) are sampled from information sets that did not end up with win of the corresponding player in the game records.

The next parameters to be described are those related to the lose scenario.
These parameters, such as $P(a_{\text{TakeWin}}\text{from} \ i | u_{0}, a)$ in Eq.~(\ref{eq:value-dahai}), can be determined by $P^j_\text{lose}(h, \phi_\text{hand})$, the probability that another player $j$ chooses $a_\text{TakeWin}$ when $i$ discards $h$ in $u_{0}$ and the hand ends immediately with game situation $\phi_\text{hand}$.
Because $j$'s hand must be tenpai when $j$ chooses $a_\text{TakeWin}$, the probability can be factorized as
\begin{eqnarray}
P^{j}_{\mathrm{lose}}(h, \phi_{\text{hand}}) = P^{j}(h, \phi_{\text{hand}}|j \text{ is tenpai}) p^{j}_{\text{tenpai}}.
\end{eqnarray}
We infer the conditional probability $P^{j}(h, \phi_{\text{hand}}|j \text{ is tenpai})$ in two different ways.
When $j$ has chosen no action in $A_{\text{Take\&Discard}}$ since $A_{\text{HandDistribution}}$, it is inferred in such a way as to further factorize the probability and draw histograms from game records.
When $j$ has chosen one or more actions in $A_{\text{Take\&Discard}}$, it is inferred in such a way as to enumerate all possible tenpai hands for $j$.
When a player has chosen actions in $A_{\text{Take\&Discard}}$ twice, the number of possible tenpai hands is order of 100 thousands, and enumerating all of them does not significantly affect the total calculation time.
When the number of such actions that player $j$ has chosen is one, it is not realistic to enumerate all tenpai hands.
However, it is possible to enumerate the remaining seven tiles by ignoring one mentsu.

\subsection{Outline of Search Algorithm of MDPs}
\label{subsec:search}

Our search algorithm to compute the expected final rank of a player at an information set has computational complexity proportional to the number of states of $\cal{M}$.
Even ignoring actions in $a_{\text{Take\&Discard}}$, there are about $10^{11}$ patterns of a player's hand, and it is not realistic to search all states related to each hand.
It is therefore desirable to reduce a sufficient number of states and actions of $\cal{M}$ so that the search algorithm ends with a realistic computational resource and the error of expected final rank does not increase.

For the purpose of such reductions, we focus on states and actions related only to hands that can realize tenpai with a relatively small number of tile exchange.
We construct a set of such hands by carrying out the following four steps:
(1) consider a graph where a vertex represents a hand, an edge represents a tile exchange, and the graph takes into account all possible hands and tile exchanges,
(2) enumerate paths with length $n$ or less connecting the current hand $q_{0}$ and a tenpai hand,
(3) construct the set of hands $Q_{\text{S0}}(n)$ by enumerating vertices along all the paths including two terminals (i.e., $q_{0}$ and a tenpai hand), and
(4) construct the set of hands $Q_{\text{S}}(n,m)$ consisting of all hands $q$ satisfying the condition that $q'$ in $Q_{\text{S0}}(n)$ exists such that $m$ or fewer mentsus of $q$ are revealed by taking from $q$.

Two integers $n$ and $m$ are parameters that control the size of the search space.
Parameter $n$ must be greater than or equal to the shanten-number of $q_{0}$ because the space must have some tenpai hands.
As $n$ and $m$ are larger, the final rank prediction is expected to be more accurate.
In our experiment, we adjusted these parameters according to the shanten-number of $q_{0}$ so that AI player can make each decision in a few seconds with light-weight desktop computers.
In this way, the size of $Q_{\text{S}}(n,m)$ is controlled to be about 50,000.
The search algorithm ignores any action that realizes a hand not belonging to $Q_{\text{S}}(n,m)$.
Our search algorithm is based on retrograde analysis \cite{Schaeffer2}, where the state values are determined from states with larger  $t$.

\subsection{Dealing with Some Popular Rules}
\label{subsec:other}
In this section, we describe how our AI player deals with some popular rules.
Dora is a tile that increases the points of a hand if it is in the winning hand.
The Dora tile is selected by a dora indicator tile, which is chosen by the chance player with $A_{\text{HandDistribution}}$.
This choice is shared by all players.
The payoff of win or lose is determined in accordance with the dora tiles.

Riich declaration is an action that can be chosen by a player who formed a tenpai hand without choosing an action in $A_{\text{Take\&Discard}}$ since $A_{\text{HandDistribution}}$.
The player who declared riich is unable to change hands but is able to earn more points when he/she wins.
We deal with riich declarations by adding hands after the declaration in $Q_{\text{S}}(n,m)$ and modifying the payoff of win $U_{\text{DrawWin}}(q, h)$ if $q$ is the hand after the declaration.
In addition, the folding tendency, i.e., other players tend to fold the hand when one declares riich, is reflected by modifying the values of $P(\overline{a_{\text{OtherWin}}} | q,t)$ and $P_{T}(h | q,t)$ according to $q$.


\section{Experiments}
\label{sec:experiment}

This section presents the results of gameplays vis-\`{a}-vis existing AI players.
We constructed the AI player with the proposed method as follows.
When the shanten-number of the hand is zero or one, we use Eq. (\ref{eq:value-inference1}) to evaluate the values of legal actions.
When the shanten-number of the hand is two or three, we use Eq. (\ref{eq:state_value_ml}) to evaluate these values.
In both cases, the player is greedy, i.e. the action with the highest value was selected.
We tentatively set $\alpha = 0.1$ in Eq. (\ref{eq:MDP-fold}).
When the shanten-number of the hand is greater than three, we adopt a simple rule-based strategy.
The rules used in this strategy basically determine whether to decrease shanten number to win or fold current hand.
To decrease the shanten-number, the rules state to choose one of the isolated tiles to discard.
To fold the hand, the rules state to choose a tile on the basis of value estimation using $\cal{M}_{\text{fold}}$.
The three AI players are one Bakuuchi and two copies of manue.
The version of Bakuuchi is the one that achieved its highest grade and ratings (R2206) in tenhou\footnote{http://tenhou.net/}, and is stronger than that published in a previous paper \cite{Mizukami}.
Table. \ref{tab:experiment} lists the result from 3557 gameplays of mahjong with the tonpu rule\footnote{This took several months using an ordinary desktop PC.}.
Because manue is clearly week, we pay attention to the difference between two ranks of our AI player and Bakuuchi for each gameplay, and observed that the mean and deviation of the difference are 0.0574 and 1.822, respectively.
Given the sample size was $3.56 \times 10^3$, the sample mean was $5.74 \times 10^{-2}$, and the sample standard deviation was 1.822, the mean was positive with one-tailed significance level $0.03$ from the analysis using the standard error of the mean.
This indicates that the performance of the AI player constructed with the proposed method reached the world heighest level.

\begin{table}[htb]
\centering
\caption{
Experimental results of 3557 gameplays of mahjong with tonpu rule.
1st to 4th columns show emperical probability obtained from results corresponding to each final ranking.
}
\begin{tabular}{|c|c|c|c|c|c|} \hline
& 1st & 2nd & 3rd & 4th & Average Ranking\\ \hline
Our AI player & 0.33 & 0.28 & 0.21 & 0.17 & 2.23 $\pm$ 0.04 \\ \hline
Bakuuchi & 0.32 & 0.27 & 0.21 & 0.20 & 2.29 $\pm$ 0.04 \\ \hline
manue & 0.17 & 0.22 & 0.29 & 0.31 & 2.74 $\pm$ 0.02 \\ \hline
\end{tabular}
\label{tab:experiment}
\end{table}

\section{Conclusion}
We proposed a method of building a state-of-the-art AI mahjong player.
With this method, multiple MDPs are introduced related to scenarios of a hand.
When the shanten-number of the hand is less than two, MDP $\cal{M}$ plays an essential role for estimating actions values in the original game.
It takes into account as many scenarios as possible, and the analysis results are directly used for evaluation of actions in the original game.
When the shanten-number of the hand is two or more, we use the results of $\cal{M}_{\text{win}}$, $\cal{M}_{\text{tenpai}}$, and $\cal{M}_{\text{fold}}$.
These MDPs are focused on a few specific scenarios, and the analysis results are used as features for inferring state values.
We reduced the number of MDP states to the extent that the expected final-rank error does not increase so that the calculation ends in a few seconds.

We presented the results of 3557 gameplays of mahjong with the AI player constructed with the proposed method and two current AI players, i.e., one version of Bakuuchi, the strongest player, and two versions of manue whose source code is published.
The results indicate the effectiveness of the proposed method.


%

\appendices


\section*{Acknowledgment}

The authors would like to thank Naoki Mizukami for his fruitful comments and supports for experiments.
This work was supported by JSPS KAKENHI Grant Numbers JP16K00503 and JP18H03347.

\ifCLASSOPTIONcaptionsoff
  \newpage
\fi



%

\bibliographystyle{IEEEtran}
\bibliography{IEEEabrv,citation}

%








\end{document}